\documentclass[runningheads]{llncs}
\usepackage{graphicx}
\usepackage{svg}
\usepackage{amsmath}
\usepackage{easyReview}
\usepackage{multirow}
\usepackage{array}
\newcolumntype{P}[1]{>{\centering\arraybackslash}p{#1}}
\newcolumntype{C}[1]{>{\centering\arraybackslash}m{#1}}

\begin{document}
\title{Improving Solar Flare Prediction by Time Series Outlier Detection}

% \author{Anonymous ICAISC 2022 Submission}
% \institute{}
%
%\titlerunning{Abbreviated paper title}
% If the paper title is too long for the running head, you can set
% an abbreviated paper title here
%
\author{
Junzhi Wen\thanks{Corresponding author: Junzhi Wen\\jwen6@student.gsu.edu}\orcidID{0000-0002-9176-5273}
\and Md Reazul Islam
\and Azim Ahmadzadeh\orcidID{0000-0002-1631-5336} 
\and Rafal A. Angryk\orcidID{0000-0001-9598-8207}
}
% \email{\{jwen6,mislam20\}@student.gsu.edu, aahmadzadeh1@gsu.edu}, angryk@cs.gsu.edu}

% %
% \authorrunning{F. Author et al.}
% % First names are abbreviated in the running head.
% % If there are more than two authors, 'et al.' is used.
% %
\institute{Georgia State University\\ Atlanta, GA 30302, USA
}

\maketitle              % typeset the header of the contribution
\begin{abstract}
    Solar flares not only pose risks to outer space technologies and astronauts' well being, but also cause disruptions on earth to our high-tech, interconnected infrastructure our lives highly depend on. While a number of machine-learning methods have been proposed to improve flare prediction, none of them, to the best of our knowledge, have investigated the impact of outliers on the reliability and robustness of those models' performance. In this study, we investigate the impact of outliers in a multivariate time series benchmark dataset, namely SWAN-SF, on flare prediction models, and test our hypothesis. That is, there exist outliers in SWAN-SF, removal of which enhances the performance of the prediction models on unseen datasets. We employ Isolation Forest to detect the outliers among the weaker flare instances. Several experiments are carried out using a large range of contamination rates which determine the percentage of present outliers. We assess the quality of each dataset in terms of its actual contamination using TimeSeriesSVC. In our best findings, we achieve a 279\% increase in True Skill Statistic and 68\% increase in Heidke Skill Score. The results show that overall a significant improvement can be achieved for flare prediction if outliers are detected and removed properly.

\end{abstract}
\section{Introduction}
    Solar flares are abrupt bursts of energy from the Sun that emit large amounts of electromagnetic radiation. They are frequently accompanied by a coronal mass ejection (CME), which is a huge bubble of radiation from the Sun. While most of the radiation and particles from a solar flare are filtered by the earth's atmosphere, intense solar flares still can release radiation that may penetrate and interfere with the radio communications, cause power outages, and pose irreversible health risks to astronauts engaging in extravehicular activities. Based on the peak flux of soft X-ray with a range of wavelengths from 0.1 to 0.8 nanometers detected by National Oceanic and Atmospheric Administration (NOAA)'s GOES satellites, solar flares are logarithmically classified into five classes as follows, namely A, B, C, M, and X, from weakest to strongest. An X-class flare is ten times stronger than an M-class flare, a hundred times stronger than a C-class flares, and so on. Among the five classes, M- and X-class flares are often targeted in space-weather prediction because they are much more likely to cause adverse effects to the earth.

    Due to the potential threats that solar flares pose to human society, flare prediction has been receiving a lot of attention during the past two decades. As machine learning techniques have achieved remarkable success in multiple fields over the recent years, they have been employed to predict flares as well \mbox{\cite{MASSONE2018355}}. Since solar flares are a spatiotemporal phenomenon with a pre-flare phase \cite{preflare}, it has been suggested that to achieve a higher and more robust performance, time series of predictive parameters for flare forecasting should be used rather than point-in-time values. While a number of machine learning methods using time series have been introduced to improve flaring prediction \mbox{\cite{HowTT2021,chen2021towards,hostetter2019understanding}}, none of them, to our best knowledge, has systematically investigated the impact of outliers in time series data for flare prediction.
    
    Outlier detection is an important task for many data mining and machine learning applications, and it has been applied broadly in various domains such as economy, biology, and astronomy \cite{benkabou2018unsupervised}. Outliers are the data instances that differ substantially from the rest of the data. In classification tasks, outliers can mislead the classifier resulting in poor performance. Therefore, the discovery of outliers is crucial to better understand the underlying nature of the data and develop more efficient methods. In this study, we investigate the possibility of improving the performance of flare prediction algorithms by removing outliers that are detected by an outlier detection algorithm, named Isolation Forest (iForest) \cite{iForest}, from a multivariate time series benchmark dataset, Space Weather ANalytics for Solar Flares (SWAN-SF) \cite{angryk2020multivariate}. We hypothesize that there exist outliers that negatively impact flare prediction in SWAN-SF.
    
    The rest of the paper is organized as follows: In Section 2, we introduce some advanced and popular outlier detection algorithms for time series data. In Section 3, we briefly describe the SWAN-SF dataset that we use in this study. In Section 4, we discuss our selection of the outlier detection algorithm. We also introduce the experiment design, methodology for tackling the class-imbalance issue, the classifier chosen, and the hyperparameter tuning process, as well as the metrics for evaluation. In Section 5, we talk about our experiments and discuss the results. Finally, we conclude and propose future work in Section 6.
% \vspace*{-\baselineskip}
    
\section{Related Work}
    Generally, outlier detection can be done locally and globally for time series data. The former means the detection of outliers within time series and the latter concerns the outliers among a set of time series data. In global outlier detection, different algorithms have been proposed based on the requirements in different fields and they could be supervised or unsupervised depending on the availability of labels in the data \cite{gupta2013outlier}.
    
    Unsupervised outlier detection can be achieved by discriminative methods, which rely on a similarity function that measures the similarity between two time series sequences. Once the similarity function is defined, a clustering mechanism is applied to cluster the data instances such that within-cluster similarity is maximized and between-cluster similarity is minimized. An outlier score is then assigned to each testing instance based on the distance to its closest cluster's centroid (or medoid). SequenceMiner \cite{od_airline} uses longest common subsequence (LCS) as the similarity measure to handle time series with different lengths, but the time series has to be discretized, which will cause the loss of information. In \cite{benkabou2018unsupervised}, dynamic time warping (DTW) is used to address the distortion in the time axis and calculate more accurate similarity between two time series. However, the high computational complexity makes it impossible to train a model on a large dataset in a reasonable time. 
    
    More recently, unsupervised outlier detection methods based on deep learning have received a lot of attention. The GGM-VAE \cite{pmlr-v95-guo18a} employs Gated Recurrent Unit (GRU) cells under a variational autoencoder (VAE) framework to discover the correlations among multivariate time series data. The Robust Deep Autoencoders (RDA) \cite{od_autoencoder} combines deep autoencoders and robust principal component analysis (RPCA) to isolate noise and outliers in the input data. The Multi-Scale Convolutional Recursive EncoderDecoder (MSCRED) \cite{zhang2019deep} jointly considers time dependence, noise robustness, and interpretation of anomaly severity. Although these methods obtain good results, they do not take into account the training time (or energy consumption) on large datasets. 
    
    Isolation Forest (iForest) \cite{iForest}, different from other outlier detection methods, is a tree-based algorithm that explicitly isolates outliers. Because of its capability of running fast on large and high-dimensional datasets, iForest has been utilized for multivariate time series data in a variety of domains \cite{audibert2020usad,iot}. More details about iForest are explained in Sec.~\ref{sec:iForest}.

\section{SWAN-SF Dataset}\label{sec:swan-sf}
    Space Weather ANalytics for Solar Flares (SWAN-SF) is a benchmark dataset introduced by \cite{angryk2020multivariate}, which entirely consists of multivariate time series (MVTS) data. The development of SWAN-SF provides a unified testbed for solar flare prediction algorithms. The dataset contains $4,075$ MVTS data instances from active regions of $10,000$ flare reports spanning over 8 years of solar active-region data from Solar Cycle 24 (May 2010 - December 2018). Each MVTS data instance of SWAN-SF represents a $12$-hour observation window of 51 flare-predictive parameters. Each time series has $60$ records with a $12$-minute cadence, and corresponds to a reported active region.
    
    The data instances in SWAN-SF are collected through a sliding-window methodology with a $1$-hour step size. A MVTS data instance is labeled by the class of the strongest flare reported within a $24$-hour prediction window right after the observation window. If no flare happens or only A-class flares are reported within an observation window, the data instance is labeled as a flare-quiet instance, denoted by N. In this work, we run a few experiments on a dichotomous version of SWAN-SF, as well as its original 5-class version. That is, we group the X- and M-class flares into one group and treat them as \textit{flaring} instances, and group the other classes (including the N class) and treat them as \textit{non-flaring} instances. The instances of the former group are also denoted by the XM class, and correspondingly, the instances of the latter group are denoted by the CBN class.
    
    Because of the sliding-window methodology used for the creation of SWAN-SF, caution must be taken when dealing with the \textit{temporal coherence} of data \cite{HowTT2021}, which can be briefly described as follows: since temporally adjacent time series have over $91\%$ of overlap, random sampling of data in order to create non-overlapping training, validation, and test sets, will fail. This introduces bias to the learner and also obscures possible overfitting. To properly deal with the temporal coherence, we take advantage of the fact that the dataset is already split into five non-overlapping partitions, and each partition has approximately the same number of X- and M-class flares. Therefore, the training and testing datasets in our experiments are selected from different partitions to prevent the effect of temporal coherence. The details of the sample sizes for each partition of SWAN-SF are listed in Table~\ref{tab1}. 
    
    SWAN-SF also exhibits extreme class imbalance. In each partition, the number of the flaring instances, which are referred to as the minority class, is significantly less than the instances of the non-flaring instances, which are referred to as the majority class. For instance, there is a 1:364 imbalance ratio between X and N and a 1:58 imbalance ratio between XM and BCN in partition 1. The imbalance ratio of each partition of SWAN-SF is listed in Table~\ref{tab1}. 
    \vspace*{-3mm}
    \begin{table}[t]
        \caption{The sample sizes and imbalance ratios of each partition in SWAN-SF}\label{tab1}
        \centering
        \begin{tabular}{|P{2cm}|C{1cm}|C{1cm}|C{1cm}|C{1cm}|C{1cm}|C{2cm}|C{2cm}|}
        \hline
        \multirow{2}{*}{Partition} & \multicolumn{5}{c|}{Class} & %
            \multicolumn{2}{c|}{Imbalance Ratio} \\
        \cline{2-8}
         & X & M & C & B & N & X:N & XM:CBN\\
        \hline
         Parition 1 & 165 & 1089 & 6416 & 5692 & 60130 & 1:364 & 1:58 \\
        \hline
        Parition 2 & 72 & 1392 & 8810 & 4978 & 73368 & 1:1019 & 1:62\\
        \hline
        Parition 3 & 136 & 1288 & 5639 & 685 & 34762 & 1:256 & 1:29\\
        \hline
        Parition 4 & 153 & 1012 & 5956 & 846 & 43294 & 1:283 & 1:43\\
        \hline
        Parition 5 & 19 & 971 & 5763 & 5924 & 62688 &  1:3299 & 1:75\\
        \hline
        \end{tabular}
    \end{table}
    % \vspace*{-\baselineskip} 

\section{Methodology}\label{sec:methodology}
    \subsection{Outlier Detection Algorithm}\label{sec:iForest}
        Isolation Forest (iForest) \cite{iForest} is a tree-based algorithm that detects outliers (anomalies) efficiently and effectively. Unlike most existing model-based outlier detection approaches, which construct a profile of normal instances and then identify instances that do not conform to the profile as outliers, iForest explicitly isolates outliers by random partitioning \cite{iForest}. Since anomalies consist of fewer instances and are assumed to have attribute values that are very different from normal instances, they are more likely to be isolated earlier during random partitioning (i.e., shorter paths in a tree structure).
        
        An iForest consists of a number Isolation Trees (iTrees), and each iTree is built on a random sample of data by recursively dividing it with a randomly selected attribute and a randomly selected split value until a terminating condition is satisfied. Each data instance is then passed through the iForest and receives an outlier score. A pre-defined \textit{contamination rate}, i.e., the proportion of outliers in the dataset, is required to provide the iForest algorithm with a halting criterion. Given a contamination rate $r$, the first $r$\% of the instances with higher outlier scores will be flagged as the outliers. Because of the effectiveness of iForest on large collections of high-dimensional data \cite{iForest}, it fits very well to our outlier detection investigation on SWAN-SF.

    \subsection{Experiment Design}
        The experiments in this study are generally designed to test the hypothesis that there exist outliers in the SWAN-SF dataset, which can negatively affect flare prediction. As mentioned in Sec.~\ref{sec:swan-sf}, there are 51 flare-predictive parameters in SWAN-SF. In this proof-of-concept study, we limit our experiments to a subset of five flare-predictive parameters only. These parameters are chosen from the top ranked features previously discovered in \cite{bobra2015solar}, namely (as named in the dataset) TOTUSJH, TOTBSQ, TOTPOT, TOTUSJZ, and ABSNJZH (see \cite{angryk2020multivariate} for the list of all parameters and their definitions). To minimize the learning bias, we use Partition 1 as the training set in all experiments and report the performance on the remaining partitions.
    
        Using iForest, given a contamination rate, we first detect outliers among the non-flaring instances in Partition 1. We then remove the detected outliers and combine the rest of the non-flaring instances with the flaring instances to build a flare forecasting model. We repeat this by gradually increasing the contamination rate (from 0.0 to 50.0) and producing new training datasets. In each case, we apply the \textit{climatology-preserving undersampling} (as discussed in Sec.~\ref{subsec:tackling_class_imbalance}) to mitigate the effect of the extreme class-imbalance issue on training our classifiers. Min-max normalization is followed to avoid the influence of different scales between parameters \cite{HAN201283}. To show the robustness of the model, we carry out 10-fold cross validation, i.e., we repeat the undersampling and normalization processes 10 times and report the average and the variance of the performance of the classifiers trained on those different datasets for each contamination rate.  
        
\begin{figure}
            \centering
            \includegraphics[width=\textwidth]{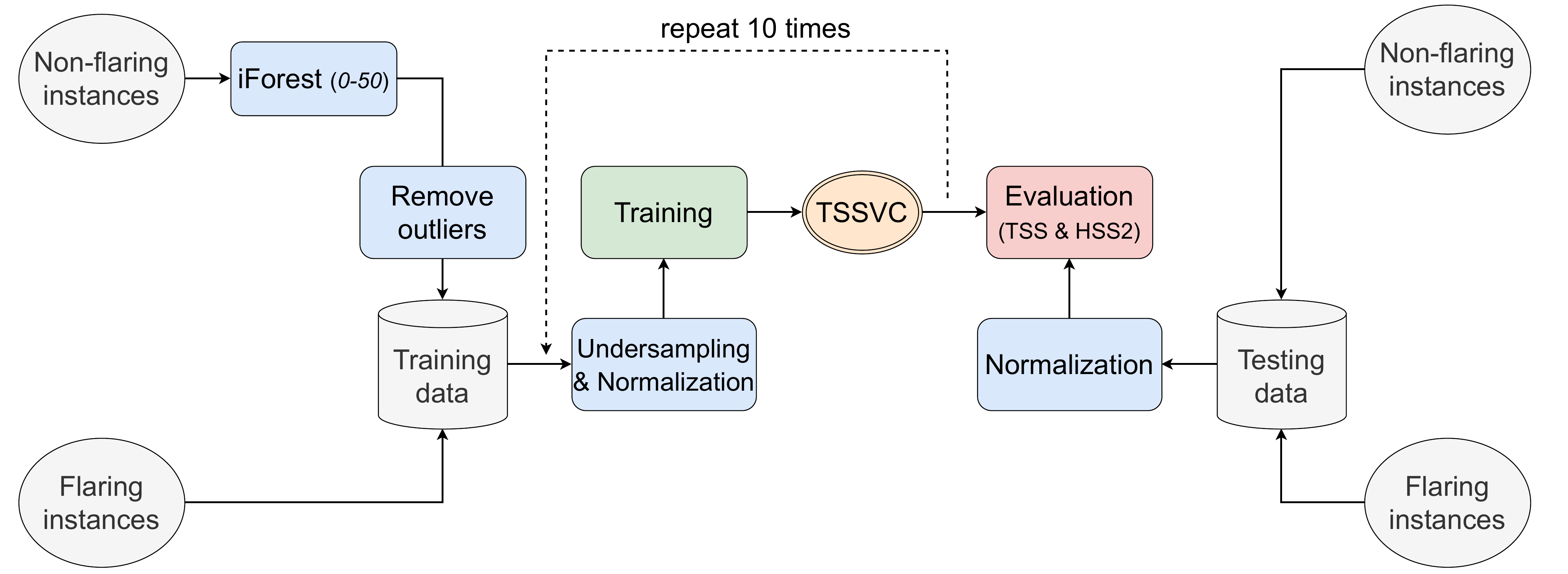}
            \caption{Experiment Design: the outlier detection operation is only applied to the non-flaring instances in Partition 1. Only normalization is applied to the test sets. 10-fold cross validation is applied for a robust evaluation. Note that the flaring instances and non-flaring instances used in Sec.~\ref{subsec:experiment_a} and Sec.~\ref{subsec:experiment_b} are different.}
            \label{fig:experiment_design}
        \end{figure}

        The trained models are tested on Partitions 2 through 5 separately. The test sets are kept unchanged except for normalization; no outlier detection, outlier removal, or undersampling is applied. This strategy provides a series of unbiased experiments, which is closest to the operational setting. The design of our experiments is shown in Fig.~\ref{fig:experiment_design}.
        
    % \vspace*{-\baselineskip} 
    \subsection{Tackling the Class-Imbalance Issue}\label{subsec:tackling_class_imbalance}
        The class-imbalance issue can affect the performance of any classifier resulting in superficially good classification/prediction scores, as thoroughly discussed in \cite{HowTT2021}. We use the so-called climatology-preserving undersampling, as suggested in \cite{HowTT2021}, which achieves a 1:1 ratio between the minority and majority instances while preserving the distribution of flare classes. Preserving the distribution of flare classes is important during undersampling because it produces more realistic data for training the model, which will hence generate a reliable model performance on testing data. Although more advanced strategies, like synthetic data generation, can be used to handle the class-imbalance issue, to avoid confounding variables that we cannot control, climatology-preserving undersampling is preferred in our study.

    \subsection{Evaluation Model and Hyperparameter Tuning}
       To evaluate the performance of our binary classification task, we choose Support Vector Machines (SVM) as the classifier. However, since we use multivariate time series data, regular SVM is not appropriate for our experiments. TimeSeriesSVC, from the tslearn machine learning toolkit \cite{tssvc}, is the SVM classifier designed specifically for time series data and is employed as the classifier in this study. TimeSeriesSVC operates support vector classification by casting DTW distances measure as definite kernels for time series \cite{cuturi2011fast}.
       
       In order to achieve an optimal performance for TimeSeriesSVC on SWAN-SF, we use the exhaustive grid-search method to tune models' hyperparameters. Ideally a grid search should be carried out separately using each dataset to find one optimal model per contamination rate. However, although such an independent optimization may result in higher classification performance, it may not necessarily bring robustness; in operational settings (i.e., real-time flare forecasting) the information about which subset of data works best for a trained model is unknown. More importantly, note that our objective is to present a fair comparison between all such models with respect to their unique contamination rates. Data-specific tuning introduces a confounding factor that we cannot control and therefore, it results in experimentation bias. Because of these reasons, we only apply grid search on the classifier that is trained on the dataset without removing any outliers, and use the optimal hyperparameters to train models across all other datasets obtained by outlier removal using different contamination rates.

    \subsection{Evaluation Metrics}
        Many measures have been developed for evaluation of the deterministic performance of classifiers using the four quantities of the confusion matrix \cite{confusionmatrix}: true positives (TP), true negatives (TN), false positives (FP), and false negatives (FN). In flare prediction studies, the True Skill Statistic (TSS) \cite{tss} and the updated Heidke Skill Score (HSS2) \cite{hss2} are typically used for performance evaluation (e.g., in \cite{benvenuto2018hybrid,Bloomfield_2012,bobra2015solar,jolliffe2012forecast,sadykov2017relationships}), and they are used in this study as well. Next, we briefly review these measures and the reasons justifying their appropriateness for a rare-event classification problem such as flare forecasting.
        
        TSS, as shown in Eq.~\ref{eq:tss}, measures the difference between the probability of true prediction (i.e., true positive rate) and the probability of false alarm (i.e., false positive rate). TSS ranges from -1 to 1, where -1 indicates that every prediction the classifier makes is incorrect, and 1 indicates a perfect performance meaning the classifier is correct for all of its predictions.  
        \begin{equation}
            TSS = \frac{TP}{TP+FN} - \frac{FP}{FP+TN}
            \label{eq:tss}
        \end{equation}

        However, the drawback of TSS is that it equates all models for which the difference between the true-positive and false-positive rates is the same. This is not a universally sound assumption (see the numerical examples in \cite{HowTT2021}). Therefore, it might be misleading to use TSS alone. This is why it is coupled with HSS2. As shown in Eq.~\ref{eq:hss}, HSS2 measures the fractional improvement of prediction that the classifier has over a random guess (no-skill) model. Similar to TSS, HSS2 ranges from -1 to 1, with 1 indicating a perfect performance, -1 indicating reverse assignment of labels to all instances, and 0 indicating no skill (i.e., as same as a random guess). 
        \begin{equation}
            HSS2 = \frac{2((TP\cdot TN) - (FN\cdot FP))}{P(FN+TN)+N(TP+FP)}
            \label{eq:hss}
        \end{equation}

\section{Experiments, Results, and Discussion} 
    \subsection{Experiment A: Impact of Outliers on X-N Classification}\label{subsec:experiment_a}
        We start by simplifying the flare prediction problem where only X-class and N-class instances are used. Since they are the most extreme classes (i.e., X is the strongest and N is the weakest), the data points are far apart, and it is easy for the classifier to distinguish between them. Through this experiment, we investigate the impact of outliers on the simplest case. Outlier detection is applied to N-class instances, which are the non-flaring instances used in this case. Random undersampling is applied in this experiment since there is only one class in the majority. The hyperparameters tuned to be used for the TimeSeriesSVC classifier are RBF kernel with the coefficient $\gamma$ being 0.01 and a soft margin constant $C$ of 100.
        
\begin{figure}
            \centering
            \includegraphics[width=\textwidth]{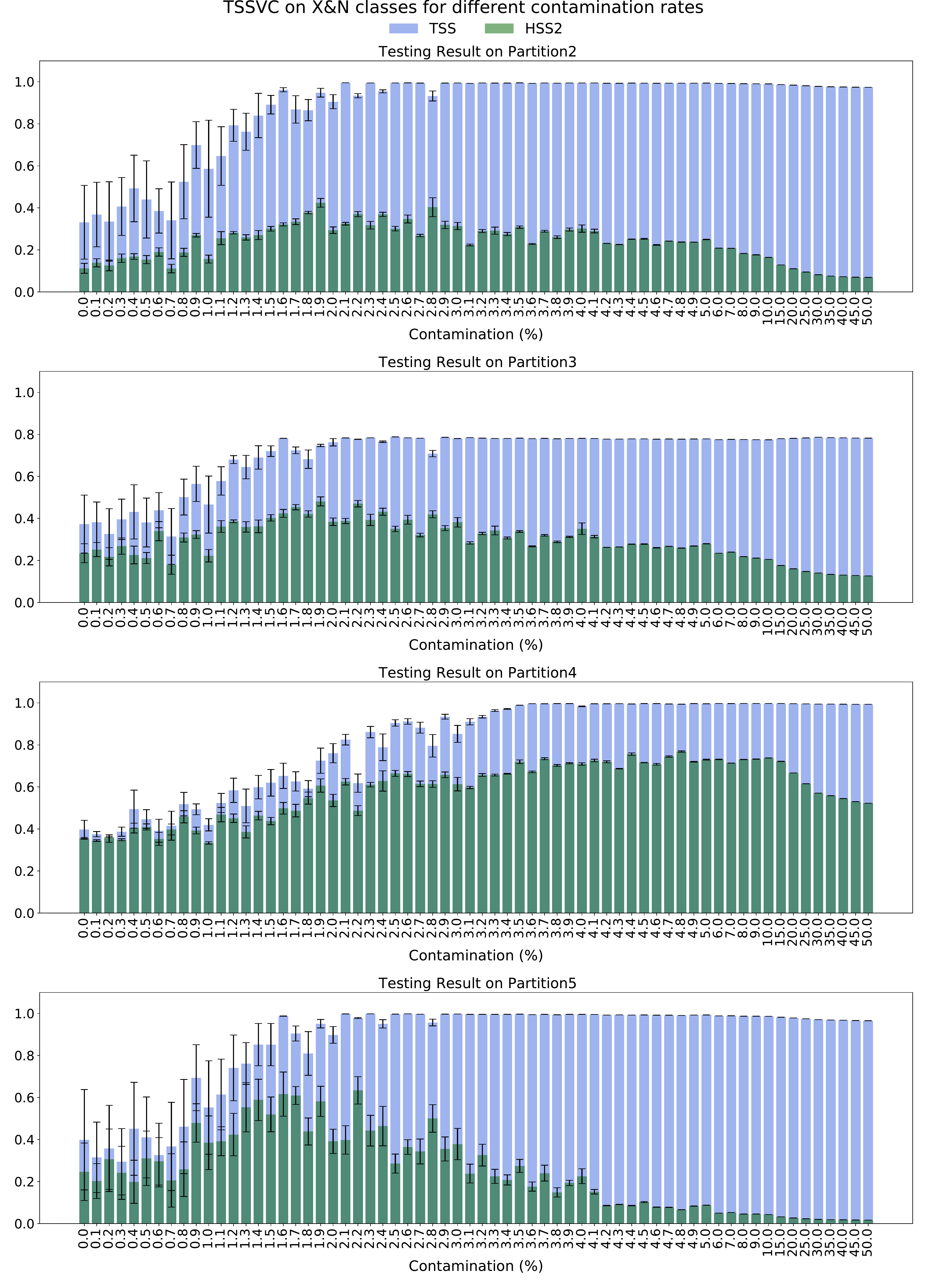}
            \caption{Result of Experiment A. The blue bar represents TSS value and the green bar represents HSS2 value at each contamination rate. The height and the black error bar of each bar represents the mean value and the variance, respectively, of the corresponding measure over 10-fold cross validation. The contamination of 0.0\% means no outlier detection applied and is the baseline.}
            \label{fig:experiment_a_barplot}
        \end{figure}

        As we can see in Fig.~\ref{fig:experiment_a_barplot}, a significant improvement is achieved by the removal of outliers. Both TSS and HSS2 increase as the contamination rate increases in the early phase, and the classifier becomes more and more robust (i.e., smaller variance). After a certain contamination rate which is unique to each partition of SWAN-SF, HSS2 starts dropping while TSS remains on the same level. Our empirical investigation shows that because at earlier stages, when the contamination rate is low, the outliers that confuse the classifier are detected and removed from the training set, hence resulting in an improvement on TSS and HSS2. However, after a certain contamination rate, iForest is forced to detect normal instances as outliers. This makes the decision boundary of the TimeSeriesSVC classifier move further towards the majority instances, so there are more FP and fewer TN, hence a smaller HSS2. 
        
    \subsection{Experiment B: Impact of Outliers on XM-CBN Classification}\label{subsec:experiment_b}
        In this experiment, we have a more complex and realistic case where the data instances are composed of all five classes of SWAN-SF, as opposed to the X- versus N-class flares that we investigated before. The non-flaring instances in this experiment consist of all instances from C, B, and N classes. Outlier detection is then applied to the group of these three classes. In addition, climatology-preserving undersampling is applied to preserve the portions of subclasses in the majority. The hyperparameters used for the TimeSeriesSVC classifier in this experiment are the same as those in Experiment A.
        
\begin{figure}
            \centering
            \includegraphics[width=\textwidth]{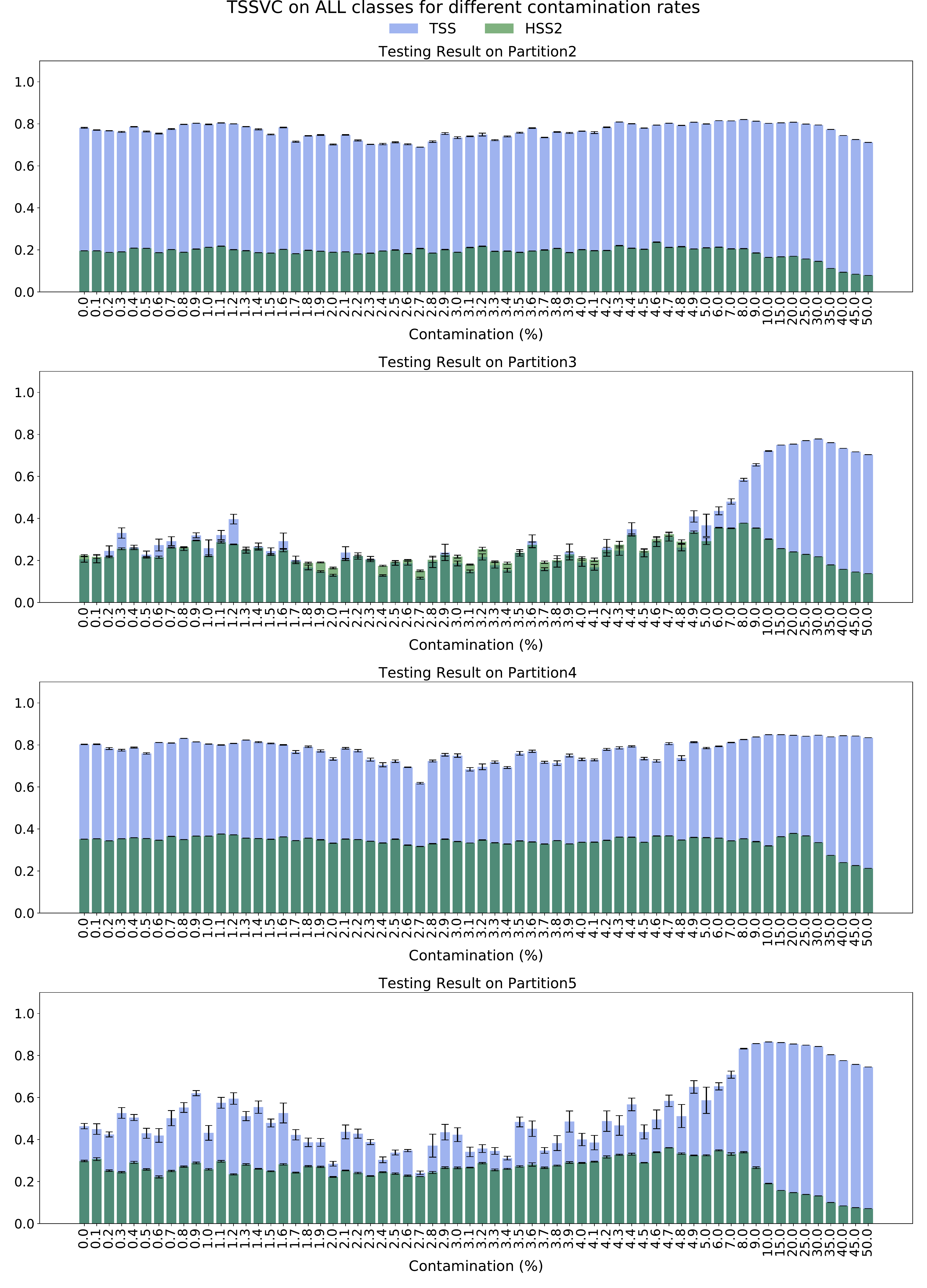}
            \caption{Result of Experiment B. The blue bar represents TSS value and the green bar represents HSS2 value at each contamination rate. The height and the black error bar of each bar represents the mean value and the variance, respectively, of the corresponding measure over 10-fold cross validation. The contamination of 0.0\% means no outlier detection applied and is the baseline.}
            \label{fig:experiment_b_barplot}
        \end{figure}

        As Fig.~\ref{fig:experiment_b_barplot} shows, there is a significant improvement of TSS on Partitions 3 and 5 while the same level of HSS2 is preserved. A minor improvement is also achieved on Partitions 2 and 4. There is a drop in terms of HSS2 at last several contamination rates on each testing partition, which is expected because too many normal data instances are forced to be removed, and this makes the decision boundary move to negative class, causing more FP, hence a smaller HSS2. Furthermore, the best TSS remains around 0.8 on each testing partition, and this may be because there also exist some outliers in flaring instances in our training data. Since, in our experiments, we only apply outlier detection to non-flaring instances (in order to preserve as many flaring instances as possible), the outliers in flaring instances are not removed. This causes more FN, hence a smaller TSS. Nevertheless, we still see that improvement can be achieved even in this simplified case.
        
\section{Conclusion and Future Work}
    In this study, we used the SWAN-SF benchmark dataset to investigate the impact of outliers in time series data. We designed two experiments to investigate how outliers affect flare prediction. After removing the outliers detected by iForest, we observed a significant improvement in the performance of the classifiers based on TSS and HSS2. This verified our hypothesis, which is there exist outliers that can negatively affect flare prediction in SWAN-SF. There are several avenues we can further explore. For example, if we can achieve a larger improvement, in Experiment B, by doing outlier detection on each individual class of non-flaring instances. Additionally, a more advanced outlier detection algorithm for multivariate time series can be utilized.

\section*{Acknowledgement}
\vspace*{-1.5mm} 
    This project has been supported in part by funding from CISE, MPS and GEO Directorates under NSF award \verb|#|1931555, and by funding from the LWS Program, under NASA award \verb|#|80NSSC20K1352.

%
% ---- Bibliography ----
%
% BibTeX users should specify bibliography style 'splncs04'.
% References will then be sorted and formatted in the correct style.
%
\bibliographystyle{splncs04}
\bibliography{mybib}
%
% \begin{thebibliography}{8}
% \bibitem{ref_article1}
% Author, F.: Article title. Journal \textbf{2}(5), 99--110 (2016)

% \bibitem{ref_lncs1}
% Author, F., Author, S.: Title of a proceedings paper. In: Editor,
% F., Editor, S. (eds.) CONFERENCE 2016, LNCS, vol. 9999, pp. 1--13.
% Springer, Heidelberg (2016). \doi{10.10007/1234567890}

% \bibitem{ref_book1}
% Author, F., Author, S., Author, T.: Book title. 2nd edn. Publisher,
% Location (1999)

% \bibitem{ref_proc1}
% Author, A.-B.: Contribution title. In: 9th International Proceedings
% on Proceedings, pp. 1--2. Publisher, Location (2010)

% \bibitem{ref_url1}
% LNCS Homepage, \url{http://www.springer.com/lncs}. Last accessed 4
% Oct 2017
% \end{thebibliography}
\end{document}